\documentclass[10pt,twocolumn,letterpaper]{article}

\usepackage{cvpr}              

\usepackage{lineno}

\definecolor{darkblue}{rgb}{0, 0, 0.5}

\usepackage[utf8]{inputenc} 
\usepackage[T1]{fontenc}    
\usepackage{url}            
\usepackage{booktabs}       
\usepackage{nicefrac}       
\usepackage{microtype}      
\usepackage{xcolor}         
\usepackage{amsfonts}       
\usepackage{amsmath}
\usepackage{caption}
\usepackage{multirow}
\usepackage{adjustbox}
\usepackage{colortbl}
\usepackage{array}
\usepackage{color}
\usepackage{graphicx}  
\usepackage{bbding}
\usepackage{amssymb}
\usepackage{pifont} 
\usepackage{wrapfig}
\usepackage{tcolorbox}
\usepackage{fontawesome5}
\usepackage{adjustbox}
\usepackage{pifont}

\definecolor{blgrey}{rgb}{0.6,0.6,0.6}
\definecolor{bblue}{rgb}{0.855,0.933,0.98}
\definecolor{dblue}{HTML}{5297D6}
\definecolor{gainred}{rgb}{0.1,0.5,0.3}
\definecolor{citecolor}{HTML}{0071BC}
\definecolor{linkcolor}{HTML}{ED1C24}
\definecolor{dkgreen}{rgb}{0,0.6,0}
\definecolor{gray}{rgb}{0.5,0.5,0.5}
\definecolor{mauve}{rgb}{0.58,0,0.82}
\usepackage{listings}
\usepackage{caption}
\usepackage{wrapfig}
\usepackage{algorithm}
\usepackage{algpseudocode} 
\usepackage{float}
\usepackage{lipsum} 
\usepackage{tabularx}
\usepackage{enumitem}  %
\usepackage{makecell}
\usepackage{pifont} %
\usepackage{algpseudocode}
\usepackage{algorithm}
\usepackage{multirow}
\usepackage{xcolor}
\usepackage{booktabs}

\definecolor{cvprblue}{rgb}{0.21,0.49,0.74}
\usepackage[pagebackref,breaklinks,colorlinks,allcolors=cvprblue]{hyperref}

\newcommand{\cmark}{\ding{51}}                           
\newcommand{\xmark}{\textcolor{lightgray!150}{\ding{55}}} 

\def\paperID{*****} 
\def\confName{CVPR}
\def\confYear{2026}

\title{Learning Geometric Representations from Videos for Spatial Intelligent\\ Multimodal Large Language Models}

\author{Haibo Wang\\
University of California, Davis\\
{\tt\small hibwang@ucdavis.edu}
\and
Lifu Huang\\
University of California, Davis\\
{\tt\small lfuhuang@ucdavis.edu}
}

\begin{document}
\maketitle
\begin{abstract}
Multimodal Large Language Models (MLLMs) excel at 2D semantic understanding but lack intrinsic 3D awareness, resulting in representations that fail to maintain geometric and spatial consistency across video frames. Given the scarcity of large-scale 3D data, we present \textbf{GeoVR}, a novel framework that learns geometric representations using purely 2D video sequences. This approach effectively restructures the semantic latent space within MLLMs to unlock spatial intelligence. Rather than employing superficial feature mixing, GeoVR reshapes the internal representations of the MLLM by distilling geometry knowledge from pre-trained 3D foundation models. This is accomplished through a multi-objective learning strategy driven by four complementary geometric targets: (1) estimating inter-frame camera poses to embed varying viewpoint dynamics, (2) regressing dense depth maps to anchor physical distances, (3) predicting a metric scale factor for real-world calibration, and (4) distilling multi-scale 3D features to align the intermediate feature space. Guided by these explicit physical and geometric constraints, the model's internal representations naturally develop strong 3D awareness. Extensive experiments on spatial reasoning benchmarks demonstrate that GeoVR achieves state-of-the-art performance, establishing a new paradigm for endowing foundation models with spatial intelligence. Code will be available at \url{https://github.com/WHB139426/GeoVR-MLLM}.
\end{abstract}    
\vspace{-10pt}
\section{Introduction}
\label{sec:intro}

\begin{figure}[t]
    \centering
    \includegraphics[width=1\linewidth]{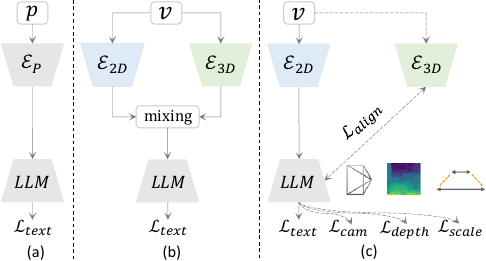}
    \caption{\textbf{Comparison of different paradigms.} $\mathcal{P}$ and $\mathcal{V}$ denote point clouds and RGB video. $\mathcal{E}_{P}$, $\mathcal{E}_{2D}$, and $\mathcal{E}_{3D}$ denote point cloud, 2D vision, and 3D foundation encoders, respectively. \textbf{(a)} relies on scarce 3D data, limiting scalability. \textbf{(b)} patches external 3D features onto 2D tokens, causing inference overhead. \textbf{(c) (ours)} restructures the latent space via training-only geometric constraints. }
    \label{fig:intro}
    \vspace{-10pt}
\end{figure}

Multimodal Large Language Models (MLLMs) \cite{Qwen3-VL, li2024llavaonevision, wang2025internvl3_5, hurst2024gpt4o, comanici2025gemini2_5} have achieved unprecedented success in 2D visual understanding tasks \cite{videomme, mvbench, wu2024longvideobench, zhou2024mlvu}. However, when deployed in scenarios involving dynamic viewpoint shifts or physical world reasoning, they often exhibit surprising brittleness \cite{yang2025vsibench, yin2025mindcube}. We attribute this vulnerability to a fundamental representation deficiency. The physical world is inherently three-dimensional, with videos acting as a dynamic projection of a consistent, implicit 3D scene under varying camera poses. However, current MLLMs are pretrained exclusively on 2D images/videos with only language supervision \cite{xu2025slowfast1.5, wang2025streambridge, llava178k, yang2025cambrians}, and their latent spaces are optimized purely for semantic alignment, ignoring the construction of intrinsic geometric representations of physical entities. Blind to physical concepts like poses, depth, and scale, these models fail to infer the implicit 3D scene.

To mitigate this issue, existing efforts generally fall into two categories. The first attempts to directly learn 3D representations by aligning LLMs with expensive and scarce explicit 3D data (e.g., point clouds), as illustrated in Figure~\ref{fig:intro} (a) \citep{hong20233dllm, xu2024pointllm, mao2025spatiallm}. However, this heavy reliance on 3D annotations severely limits data scalability and compromises the model's generalization capabilities for standard 2D visual understanding. The second approach, shown in Figure~\ref{fig:intro} (b), incorporates pre-trained 3D foundation models $\mathcal{E}_{3D}$ \cite{wang2025vggt, wang2024dust3r, lin2025depthanything3} into the MLLM architecture to supply auxiliary 3D representations. Despite the rich 3D priors encapsulated in these models, their integration is largely confined to superficial feature mixing, such as element-wise addition (e.g., VG-LLM \cite{zheng2025vgllm}, Spatial-MLLM \cite{wu2025spatialmllm}) or attention-based fusion (e.g., VLM-3R \cite{fan2025vlm3r}, SpaceMind \cite{zhao2025spacemind}). Such shallow alignment fails to fundamentally instill geometric awareness into the MLLM's intrinsic visual representations. Instead, it merely fuses the 2D tokens with external 3D features with a dual-branch architecture, thereby introducing substantial computational overhead during inference.

In contrast to these paradigms, as in Figure~\ref{fig:intro} (c), we propose \textbf{GeoVR}, a novel framework that learns geometric representations directly from pure 2D video sequences, entirely eliminating the reliance on any manual 3D annotations. Rather than superficially mixing external features, the core philosophy of GeoVR is to fundamentally restructure the MLLM's internal semantic space. We achieve this through a multi-objective learning strategy that leverages the robust geometric priors of existing 3D foundation models, not as external plug-ins, but as targets to rewire the visual tokens intrinsically. Specifically, GeoVR imposes four complementary geometric constraints exclusively during the training phase: (1) \textbf{Camera Pose Estimation}, which captures the physical logic of varying viewpoints across continuous video frames; (2) \textbf{Depth Map Prediction}, which grounds the 2D tokens with depth information, enabling the model to perceive physical distances and occlusions; (3) \textbf{Metric Scale Calibration}, which anchors the spatial features into the real-world scale, empowering the model to comprehend the absolute magnitude of the scene; and (4) \textbf{Multi-scale Geometric Representation Alignment}, which aligns the MLLM's internal latent space with the structured geometric representations of a pre-trained 3D foundation model \cite{wang2025vggt, lin2025depthanything3, wang2026vggtomega}. By confining all these explicit geometric regularizations to the training stage, GeoVR natively awakens the MLLM's 3D reasoning capabilities without introducing additional computational burden during inference.


In summary, we conclude our contributions as follows: (1) We propose \textbf{GeoVR}, a novel paradigm to restructure MLLM's intrinsic representations with geometric awareness using purely 2D videos, bypassing the scalability limits of explicit 3D annotations. (2) We design a multi-objective learning framework comprising pose estimation, depth prediction, metric scale calibration, and representation alignment, which successfully distills the multi-view geometric priors into the MLLM's latent space without additional computational overhead during inference. (3) Through extensive experiments and analysis, we demonstrate that GeoVR achieves state-of-the-art performance on spatial reasoning benchmarks.
\section{Related Work}
\label{sec:related}

\textbf{MLLMs for 3D Scene Understanding} 
has attracted significant interest recently, aiming to unify 3D understanding and visual-language reasoning. Early works rely on explicit 3D inputs. Methods such as PointLLM \cite{xu2024pointllm}, 3D-LLM \cite{hong20233dllm}, Spatial-LM \cite{mao2025spatiallm}, and LL3DA \cite{chen2024ll3da} ingest explicit 3D data (e.g., point clouds or reconstructed meshes), process them via specialized 3D encoders, and project them into the MLLM's embedding space. While effective for 3D-centric tasks, these approaches face the bottlenecks of severe scarcity of large-scale, high-quality 3D-text paired data. To bypass 3D data reliance, another line of work, such as SpatialVLM \cite{chen2024spatialvlm}, LLaVA-3D \cite{zhu2025llava3D}, and Video-3D-LLM \cite{zheng2025video3dllm}, attempts to solve spatial reasoning directly from 2D images/videos. However, they train the model with only semantics supervision, inherently lacking the capability to perceive true physical depth and multi-view consistency. In contrast, our approach entirely bypasses the need for 3D annotations and point cloud encoders, learning rich geometric representations directly from 2D video sequences.

\noindent \textbf{Feed-forward 3D Reconstruction} 
has emerged as a powerful paradigm, capable of jointly inferring varying 3D attributes in a single forward pass. This paradigm was pioneered by DUSt3R \cite{wang2024dust3r} for pairwise image inputs, and subsequently refined by MASt3R \cite{Leroy2024mast3r} for improved feature matching. More recently, the field has rapidly expanded to multi-view scenarios and video sequences, with architectural innovations such as VGGT \cite{wang2025vggt}, MapAnything \cite{keetha2025mapanything}, DepthAnyhing 3 \cite{lin2025depthanything3}, $\pi^3$ \cite{wang2026pi3}, and VGGT-$\Omega$ \cite{wang2026vggtomega}. These methods adopt simple and efficient end-to-end inference to predict 3D points, dense depths, and camera poses, often surpassing classical Structure-from-Motion (SfM) pipelines. However, despite their exceptional ability to extract low-level geometry, these models remain strictly focused on reconstruction. They lack linguistic interfaces and higher-level semantic reasoning capabilities. In our work, rather than using these models for standalone reconstruction, we exploit their robust geometric priors as distillation targets.
 
\begin{figure*}[t]
\begin{center}
\includegraphics[scale=0.567]{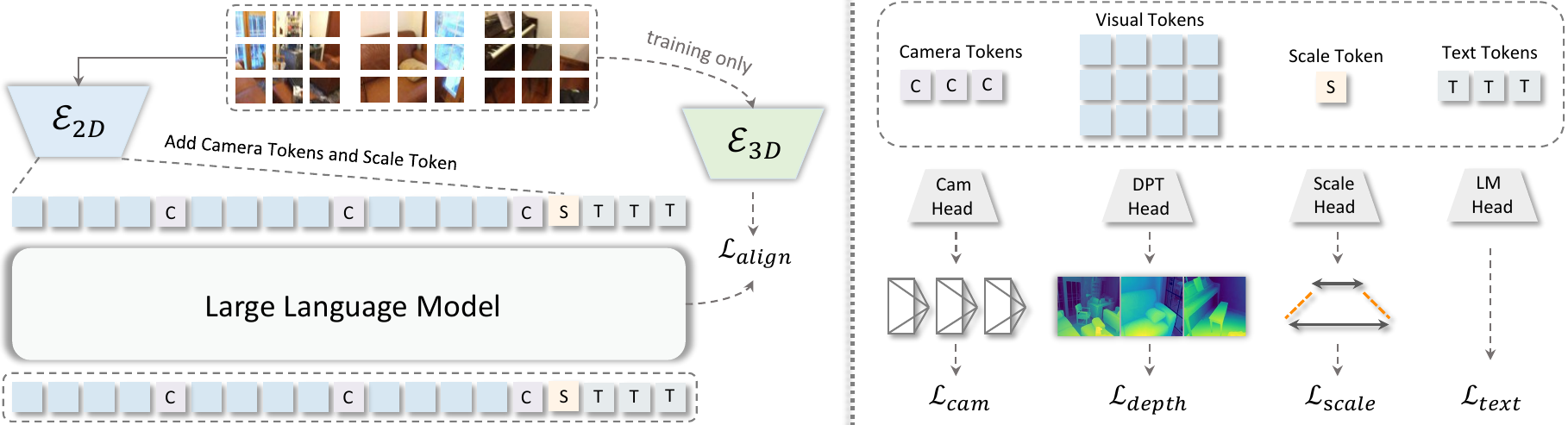}
\end{center}
\caption{\textbf{Framework of GeoVR.} During training, alongside the standard next-token prediction ($\mathcal{L}_{text}$), the MLLM representation is additionally restructured via: camera pose estimation ($\mathcal{L}_{cam}$), depth prediction ($\mathcal{L}_{depth}$), metric scale calibration ($\mathcal{L}_{scale}$), and geometric representation alignment ($\mathcal{L}_{align}$) from a frozen 3D teacher ($\mathcal{E}_{3D}$). All auxiliary heads and the $\mathcal{E}_{3D}$ branch are discarded during inference.}
\label{fig:method}
\end{figure*}

\noindent \textbf{MLLMs with 3D Foundation Models.} 
Recognizing the limitations of 2D data priors, contemporary research has begun integrating pre-trained 3D foundation models into MLLM architectures. The early approach is passive feature fusion. For instance, VG-LLM \cite{zheng2025vgllm} and Spatial-MLLM \cite{wu2025spatialmllm} extract 3D features using a frozen 3D foundation model and fuse them with 2D tokens via patch-level addition, while VLM-3R \cite{fan2025vlm3r}, SpaceMind \cite{zhao2025spacemind}, and GeoThinker \cite{li2026geothinker} inject 3D features via cross-attention. G$^2$VLM \cite{hu2025g2vlm} introduces an MoT architecture with dedicated geometric experts. However, maintaining a 3D encoder inevitably incurs a computational bottleneck during inference. There are also works such as Spatial Forcing \cite{li2026spatialforcing} and 3DRS \cite{huang20253drs} shift towards training-time alignment by distilling VGGT priors into MLLM features. Yet, these methods remain limited as they rely on singular, feature-level alignment without comprehensive physical constraints. In contrast, \textbf{GeoVR} proposes a holistic intrinsic representation restructuring. We enforce a multi-objective learning strategy strictly during training. By implicitly distilling multi-view geometry from 3D Foundation models, GeoVR endows the MLLM with profound spatial intelligence at zero additional inference cost.

\section{Method}
\label{sec:method}
We introduce \textbf{GeoVR} in Figure \ref{fig:method}, a novel framework designed to awaken spatial intelligence within MLLMs purely from 2D video sequences. The core philosophy of our approach is to fundamentally restructure the MLLM's internal semantic latent space into geometry-aware representations through multi-objective geometric learning.

\subsection{Problem Formulation}
\label{subsec:problem}
Let $\mathcal{V} = \{I_t\}_{t=1}^T \in \mathbb{R}^{T \times 3 \times H \times W}$ represent an input video comprising $T$ frames, accompanied by a text instruction $\mathcal{X}_{text}$. In the standard MLLM paradigm, a pre-trained 2D vision encoder $\mathcal{E}_{2D}$ is employed to process the sequence, extracting a set of visual tokens $\mathcal{E}_{2D}(\mathcal{V}) \in \mathbb{R}^{T \times N_{2D} \times D_{2D}}$, where $N_{2D}$ denotes the number of patch tokens per frame and $D_{2D}$ is the embedding dimension. These visual tokens are linearly projected and fed into the Large Language Model alongside the tokenized text instructions. The entire framework is conventionally optimized via the standard autoregressive next-token prediction objective:
\begin{equation}
    \mathcal{L}_{text} = - \sum_{i=1}^{L} \log P_{\theta}(y_i \mid y_{<i}, \mathcal{E}_{2D}(\mathcal{V}), \mathcal{X}_{text})
\end{equation}
where $y_i$ is the $i$-th target text token and $\theta$ is the parameters of the MLLM. However, $\mathcal{L}_{text}$ is purely language-driven supervision and lacks explicit geometric signal, bounding the internal latent space only to 2D representations, inherently collapsing the complex 3D physical world into a flat semantic space. Consequently, the resulting visual tokens fail to perceive essential geometric concepts such as pose, depth, scale, and multi-view structural consistency.

\begin{figure}[t]
    \centering
    \includegraphics[width=1\linewidth]{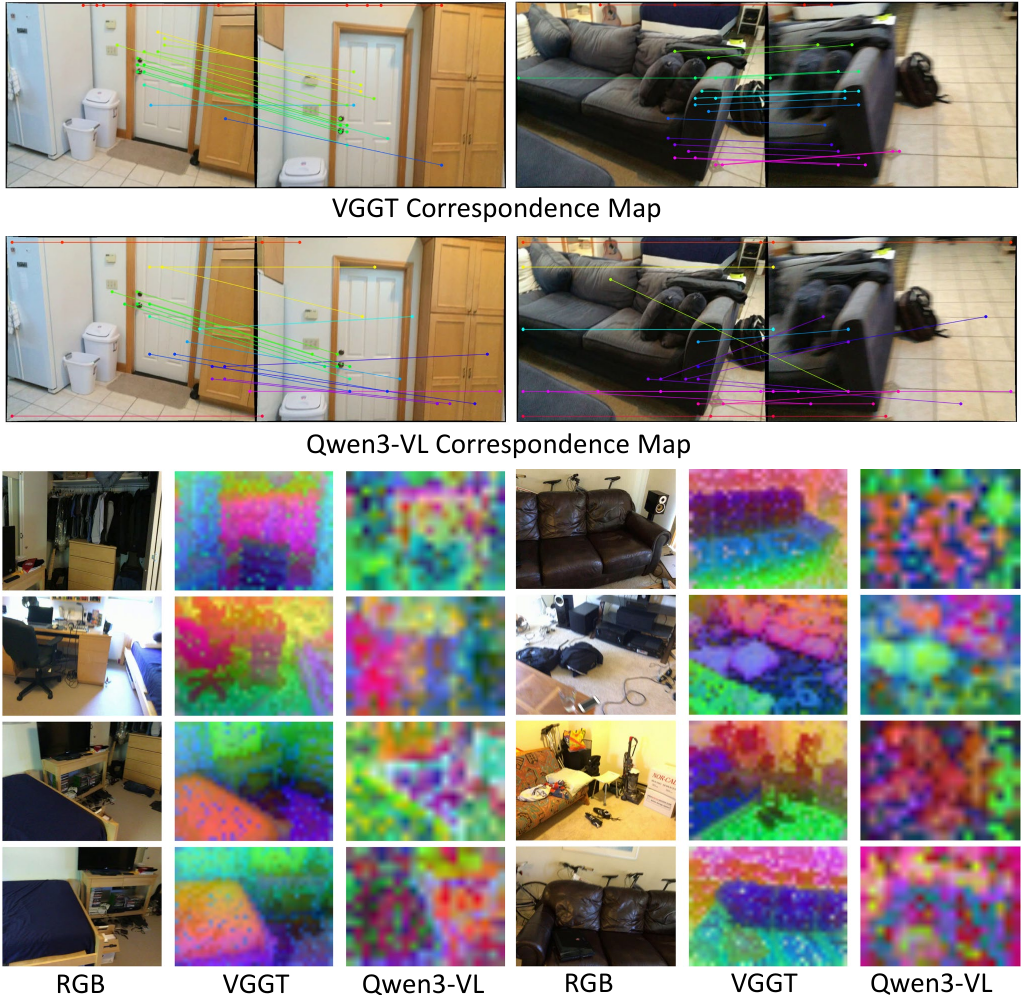}
    \caption{Cross-view correspondences and PCA projections of representations from Qwen3-VL and VGGT.}
    \label{fig:feature}
    \vspace{-10pt}
\end{figure}

To empirically validate this representation deficiency, we visualize the cross-view correspondences and Principal Component Analysis (PCA) projections of the features from an MLLM (Qwen3-VL \cite{Qwen3-VL}) against a 3D foundation model (VGGT \cite{wang2025vggt}) in Figure~\ref{fig:feature}. As illustrated, the MLLM's representations fail to establish robust correspondences across varying viewpoints and exhibit severe semantic ambiguity. For comparison, VGGT's representations accurately track physical points across the 3D scene, and maintain sharp, instance-level geometric consistency. This stark contrast empirically confirms that purely language-driven pre-training is insufficient for spatial perception, underscoring the urgent need for explicit geometric grounding.

To overcome this, we force the MLLM to reconstruct essential geometric properties using its \textit{own} representations. \textit{By optimizing for a set of geometric targets, we aim to restructure the model's internal latent space from a semantic manifold into 3D-aware representations.} Specifically, we adopt a \textit{minimalist geometric learning} strategy. By dropping heavy targets like point cloud reconstruction and tracking, we focus on four geometric targets: camera poses (Sec. \ref{subsec:camera}), depth maps (Sec. \ref{subsec:depth}), metric scale factor (Sec. \ref{subsec:scale}), and representation alignment (Sec. \ref{subsec:align}), effectively awakening the 3D awareness while preserving the model's general capacity.

\subsection{3D Foundation Model Teacher}
\label{subsec:3d_teacher}
To obtain these minimal geometric targets, we introduce a 3D foundation model (e.g., VGGT(-$\Omega$) \cite{wang2025vggt, wang2026vggtomega} or DepthAnything 3 \cite{lin2025depthanything3}) as a 3D teacher, denoted as $\mathcal{E}_{3D}$. Unlike standard 2D vision encoders, $\mathcal{E}_{3D}$ adopts a unified architecture with alternating frame-wise and global self-attention, explicitly designed to output a variety of 3D quantities directly from 2D image sequences. By feeding the same raw video $\mathcal{V} \in \mathbb{R}^{T \times 3 \times H \times W}$ into the frozen $\mathcal{E}_{3D}$, the forward pass yields several streams of geometric targets we need:

\begin{enumerate}
    \item \textbf{Camera Poses:} The camera prediction head of $\mathcal{E}_{3D}$ outputs the camera parameters (intrinsics and extrinsics) $\mathcal{P} \in \mathbb{R}^{T \times 9}$. For each frame, this 9-dimensional vector explicitly parameterizes the camera poses, comprising a 3-dimensional translation vector, a 4-dimensional rotation quaternion, and a 2-dimensional field of view.
    
    \item \textbf{Dense Depth Maps:} The depth prediction head of $\mathcal{E}_{3D}$ generates dense maps $\mathcal{D} \in \mathbb{R}^{T \times H \times W}$, associating each pixel location $(i, j)$ from the $t$-th camera frame with its corresponding depth value $\mathcal{D}_t(i, j) \in \mathbb{R}^{+}$.

    \item \textbf{Metric Scale Factor:} By aligning the up-to-scale depth maps using a Metric Depth Model \cite{lin2025depthanything3}, we derive a global metric scale factor $\mathcal{S} \in \mathbb{R}^{+}$. For each video, this scalar calibrates the relative geometric attributes (camera poses and depth maps) into absolute physical dimensions with true real-world magnitudes.

    \item \textbf{Geometric Representations:} We extract the intermediate features from multiple layers of the $\mathcal{E}_{3D}$ backbone, yielding a representation $\mathcal{F}_{3D} \in \mathbb{R}^{L_{3D} \times T \times N_{3D} \times D_{3D}}$, where $L_{3D}$ denotes the number of extracted layers. $\mathcal{F}_{3D}$ implicitly encapsulates rich geometric knowledge.
\end{enumerate}

By leveraging $\mathcal{E}_{3D}$, we dynamically generate these geometric targets $\mathcal{P}$, $\mathcal{D}$, $\mathcal{S}$ and $\mathcal{F}_{3D}$ as pseudo-labels for any arbitrary video sequence during training. This strategy decouples our GeoVR framework from the reliance on scarce, manually annotated 3D datasets. It allows our geometric representation learning to scale to large-scale, in-the-wild video corpora, bypassing the data acquisition bottleneck.
    
\subsection{Camera Pose Estimation}
\label{subsec:camera}
To natively capture the viewpoint dynamics and the observer's physical motion, we introduce a Camera Pose Estimation objective. We introduce a learnable \textbf{camera token} $\mathcal{F}_{cam} \in \mathbb{R}^{D_{2D}}$ to serve as a global receptor. For each of the $T$ frames in the video, we append $\mathcal{F}_{cam}$ to the end of its corresponding visual tokens before feeding them into the LLM. Through the deep self-attention layers, these camera tokens naturally aggregate multi-view context from the surrounding visual features across the entire video sequence.

We then extract $\mathcal{H}_{cam} \in \mathbb{R}^{T \times D_{2D}}$, corresponding to the $T$ camera tokens from the MLLM's last layer hidden states. To predict the camera state for each frame $t$, we process its corresponding hidden state $\mathcal{H}_{cam, t}$ through a lightweight \textbf{Camera Head} (a simple MLP), which regresses a 9-dimensional camera parameter vector $\hat{\mathcal{P}}_t \in \mathbb{R}^9$.

Following the 3D teacher $\mathcal{E}_{3D}$, $\hat{\mathcal{P}}_t \in \mathbb{R}^9$ is decomposed into a translation vector $\hat{\mathbf{q}}_t \in \mathbb{R}^3$, a rotation quaternion $\hat{\mathbf{t}}_t \in \mathbb{R}^4$, and a field of view vector $\hat{\mathbf{f}}_t \in \mathbb{R}^2$. 
Similarly, we denote the corresponding geometric pseudo-labels extracted from the teacher as $\mathcal{P}_t = [\mathbf{q}_t, \mathbf{t}_t, \mathbf{f}_t]$. The camera pose loss $\mathcal{L}_{cam}$ is formulated to minimize the discrepancy between the MLLM's internal predictions and the geometric pseudo-labels with a weighted $L_1$ loss:
\begin{equation}
    \mathcal{L}_{cam} = \frac{1}{T} \sum_{t=1}^{T} \left( |\mathbf{q}_t - \hat{\mathbf{q}}_t| + \beta_q |\mathbf{t}_t - \hat{\mathbf{t}}_t| + \beta_f |\mathbf{f}_t - \hat{\mathbf{f}}_t| \right)
\end{equation}
where $\beta_q$ and $\beta_f$ are factors balancing the rotation and intrinsic components. By strictly constraining these camera tokens, we compel the MLLM's attention mechanisms to implicitly capture the underlying 3D spatial transformations, effectively forcing the model to represent the video as a consistent 3D scene observed through a moving lens.

\subsection{Depth Map Prediction}
\label{subsec:depth}
To ground the visual tokens with the explicit awareness of spatial layout and physical distances, we introduce a Dense Depth Prediction objective. We extract multi-scale hidden states corresponding to the visual tokens from a selected set of layers within the MLLM to simultaneously capture low-level structural details and high-level semantic context. For each selected layer, we discard the appended camera tokens. This process yields a hierarchical feature representation $\mathcal{H}_{depth} \in \mathbb{R}^{L_{depth} \times T \times N_{2D} \times D_{2D}}$, where $L_{depth}$ denotes the number of extracted layers. This structured, multi-level feature pyramid is then fed into a lightweight Dense Prediction Transformer (DPT) Head \cite{ranftl2021dpt} (we modify some convolutional blocks with a simple MLP for efficiency). By effectively aggregating the multi-scale representations, the DPT head progressively upsamples the features to predict high-resolution dense depth maps $\hat{\mathcal{D}} \in \mathbb{R}^{T \times H \times W}$. 

To supervise this dense regression task, the depth loss uses $L_1$ loss to weigh the discrepancy between the predicted depth $\hat{\mathcal{D}}$ and the pseudo-labels $\mathcal{D}$. Following VGGT \cite{wang2025vggt}, we additionally apply a gradient-based term, which is widely used in monocular depth estimation. Therefore, the final depth loss $\mathcal{L}_{depth}$ is formulated as:
\begin{equation}
\begin{split}
    \mathcal{L}_{depth} = \frac{1}{T} \sum_{t=1}^{T} \Big(| \hat{\mathcal{D}}_t - \mathcal{D}_t | + | \nabla \hat{\mathcal{D}}_t - \nabla \mathcal{D}_t |\Big)
\end{split}
\label{equ:loss_depth}
\end{equation}
where $\nabla$ denotes the gradient operator. We discard the aleatoric uncertainty loss \cite{novotny2017uncertainty, kendall2017uncertainties} used in DUSt3R \cite{wang2024dust3r} because it can be unstable during fine-tuning \cite{wang2026vggtomega}.

\subsection{Metric Scale Calibration}
\label{subsec:scale}
 While camera pose and depth map capture the relative spatial structure and layout of the scene, monocular geometric predictions inherently suffer from scale ambiguity. To anchor these relative quantities into absolute physical dimensions, we introduce the Metric Scale Calibration objective.

Specifically, we introduce a single learnable \textbf{scale token} $\mathcal{F}_{scale} \in \mathbb{R}^{D_{2D}}$ as a video-level global aggregator, appended to the very end of the entire visual token sequence. Through the MLLM's global self-attention mechanism, it aggregates spatio-temporal geometric cues to perceive the overall magnitude of the environment. The hidden state of this token, $\mathcal{H}_{scale}$, is then processed by an MLP head with an exponential activation to regress a strictly positive absolute scale factor $\hat{\mathcal{S}} = \exp(\text{MLP}(\mathcal{H}_{scale})) \in \mathbb{R}^{+}$. We formulate the scale loss $\mathcal{L}_{scale}$ in a logarithmic space with the pseudo ground-truth scale $\mathcal{S} \in \mathbb{R}^{+}$ using an $L_1$ distance:
\begin{equation}
    \mathcal{L}_{scale} = \left| \log(1 + \hat{\mathcal{S}}) - \log(1 + \mathcal{S}) \right|
\end{equation}
This logarithmic formulation effectively compresses extreme physical dimensions, ensuring balanced gradients and stable convergence across diverse in-the-wild datasets.

\subsection{Geometric Representation Alignment}
\label{subsec:align}
\begin{figure}[t]
    \centering
    \includegraphics[width=1\linewidth]{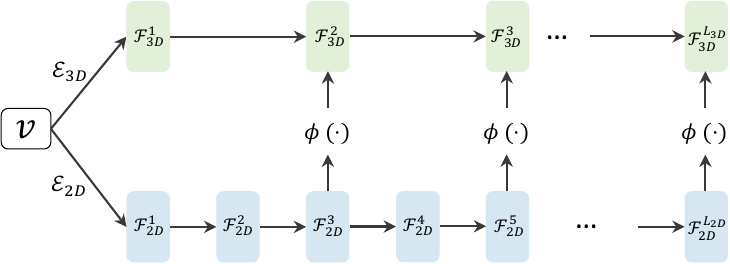}
    \caption{Distill the geometric prior from $\mathcal{F}_{3D}$ into $\mathcal{F}_{2D}$.}
    \label{fig:align}
    \vspace{-10pt}
\end{figure}

Beyond explicit targets such as camera pose and depth map, GeoVR fundamentally restructures the MLLM's representation via multi-scale distillation. As in Figure \ref{fig:align}, we align the MLLM's intrinsic latent space with the rich, structured geometric priors of the 3D foundation teacher model ($\mathcal{E}_{3D}$). Crucially, this alignment is not limited to the final output; it is enforced across multiple intermediate layers, ensuring that the MLLM develops geometric awareness at varying scales.

Formally, we extract the multi-layer hidden states $\mathcal{F}_{2D} \in \mathbb{R}^{L_{2D} \times T \times N_{2D} \times D_{2D}}$ from the MLLM, and the multi-layer geometric features $\mathcal{F}_{3D} \in \mathbb{R}^{L_{3D} \times T \times N_{3D} \times D_{3D}}$ from the 3D teacher. Here, $L_{2D}$ and $L_{3D}$ represent the total number of layers in the respective models. Due to the discrepancy in patch sizes between $\mathcal{E}_{2D}$ and $\mathcal{E}_{3D}$, the resulting token counts $N_{2D}$ and $N_{3D}$ are mismatched. To resolve this resolution gap, we introduce a projection function $\phi$, which first restores the 1D token sequence into a 2D spatial grid and applies bilinear interpolation to resize the MLLM feature maps to match $\mathcal{F}_{3D}$, followed an MLP to project the channel dimension of $\mathcal{F}_{2D}$ to $D_{3D}$. The geometric representation alignment loss $\mathcal{L}_{align}$ is then optimized by minimizing the cosine distance between the projected MLLM features and the teacher's geometric features:

\begin{equation}
\mathcal{L}_{align} = \frac{1}{|L|} \sum_{l \in L} \left( \text{Sim} \left( \mathcal{F}_{3D}^{l}, \phi \left( \mathcal{F}_{2D}^{s(l)} \right) \right) \right)
\end{equation}

where $L$ defines the set of $\mathcal{E}_{3D}$'s layer indices chosen for distillation, and $s(l)$ denotes the corresponding target layer index in the MLLM, mapped proportionally based on the model depth. $\text{Sim}(\cdot, \cdot)$ computes the cosine similarity.

\subsection{Training Objectives}
\label{subsec:train}
The overall optimization objective is formulated as a multi-task learning problem, where the model is jointly supervised by language modeling signals and explicit geometric constraints. The total loss function $\mathcal{L}_{total}$ is defined as:
\begin{equation}
\mathcal{L}_{total} = \mathcal{L}_{text} + \lambda_{1} \mathcal{L}_{cam} + \lambda_{2} \mathcal{L}_{depth} + \lambda_{3} \mathcal{L}_{scale} + \lambda_{4} \mathcal{L}_{align}
\end{equation}
where $\lambda_{1,2,3,4}$ are hyperparameters for balancing each loss term. Crucially, all auxiliary heads and the 3D teacher model are only required during training, without additional computational overhead during inference.
\definecolor{darkblue}{rgb}{0, 0, 0.5}
\definecolor{Gray}{gray}{0.93}
\definecolor{uclagold}{rgb}{1.0, 0.7, 0.0}
\definecolor{airforceblue}{rgb}{0.36, 0.54, 0.66}
\definecolor{rosegold}{rgb}{0.72, 0.43, 0.47}
\definecolor{pastelbrown}{rgb}{0.51, 0.41, 0.33}
\definecolor{isabelline}{rgb}{0.96, 0.94, 0.93}
\definecolor{macaroniandcheese}{rgb}{0.98, 0.89, 0.83}
\definecolor{wildblueyonder}{rgb}{0.85, 0.89, 0.95}
\definecolor{mediumtaupe}{rgb}{0.4, 0.3, 0.28}
\definecolor{bluegray}{rgb}{0.4, 0.6, 0.8}
\definecolor{celestialblue}{rgb}{0.29, 0.59, 0.82}
\definecolor{darkorange}{rgb}{1.0, 0.55, 0.0}
\definecolor{cadmiumred}{rgb}{0.89, 0.0, 0.13}
\definecolor{magnolia}{rgb}{0.97, 0.96, 1.0}
\definecolor{pastelblue}{rgb}{0.68, 0.78, 0.81}
\definecolor{persiangreen}{rgb}{0.0, 0.65, 0.58}
\definecolor{steelblue}{rgb}{0.27, 0.51, 0.71}
\definecolor{bluebell}{rgb}{0.64, 0.64, 0.82}
\definecolor{dimgray}{rgb}{0.41, 0.41, 0.41}
\definecolor{splashedwhite}{rgb}{1.0, 0.99, 1.0}
\definecolor{lavendergray}{rgb}{0.77, 0.76, 0.82}
\definecolor{lightgray}{rgb}{0.83, 0.83, 0.83}
\definecolor{lavendermist}{rgb}{0.9, 0.9, 0.98}
\definecolor{lightgreen}{HTML}{f8fcf4}
\definecolor{lightblue}{HTML}{dfebf7}

\definecolor{peachpuff}{HTML}{FFDAB9}      
\definecolor{mistyrose}{HTML}{FFE4E1}      
\definecolor{lavenderblush}{HTML}{FFF0F5}  
\definecolor{lemonchiffon}{HTML}{FFFACD}   
\definecolor{cornsilk}{HTML}{FFF8DC}       
\definecolor{seashell}{HTML}{FFF5EE}       

\definecolor{aliceblue}{HTML}{F0F8FF}      
\definecolor{paleblue}{rgb}{0.85, 0.92, 0.98} 
\definecolor{lightcyan}{HTML}{E0FFFF}      
\definecolor{mintcream}{HTML}{F5FFFA}      
\definecolor{honeydew}{HTML}{F0FFF0}       
\definecolor{palegreen}{rgb}{0.85, 0.96, 0.85} 

\definecolor{floralwhite}{HTML}{FFFAF0}    
\definecolor{whitesmoke}{HTML}{F5F5F5}     
\definecolor{palepurple}{rgb}{0.92, 0.88, 0.95} 
\definecolor{oatmeal}{rgb}{0.94, 0.92, 0.88}    

\definecolor{palelavender}{HTML}{F5F2F9} 
\definecolor{lilacwhite}{HTML}{F8F6FA}   
\definecolor{ghostpurple}{rgb}{0.96, 0.95, 0.98} 
\definecolor{softthistle}{rgb}{0.93, 0.91, 0.95}

\begin{table*}[!t]\large
    \centering
    \resizebox{\linewidth}{!}{
    \begin{tabular}{ c | c | c | c c c c | c c c c }
        \toprule

        \multirow{2}{*}{\textbf{Method}}
        & \multicolumn{1}{c|}{\textbf{w/o}} 
        & \multirow{2}{*}{\textbf{Avg.}} 
        & \multicolumn{4}{c|}{\cellcolor{orange!10}\textbf{Numerical Answer}} 
        & \multicolumn{4}{c}{\cellcolor{yellow!10}\textbf{Multiple-Choice Answer}} 
        \\

         \cmidrule(lr){4-7} \cmidrule(lr){8-11} 
        
         &\textbf{$\mathcal{E}_{3D}$}
         &
         &\rotatebox{0}{\textit{Obj. Count}} 
         &\rotatebox{0}{\textit{Abs. Dist}} 
         &\rotatebox{0}{\textit{Obj. Size}} 
         &\rotatebox{0}{\textit{Room Size}} 
         &\rotatebox{0}{\textit{Rel. Dis}}  
         &\rotatebox{0}{\textit{Rel. Dir}} 
         &\rotatebox{0}{\textit{Route Plan}} 
         &\rotatebox{0}{\textit{Appr. Order}}
         \\

        \midrule

        \rowcolor{gray!10!blue!5} 
        \multicolumn{11}{l}{\textit{Proprietary Models / Human}}\\

         Human 
         &- &79.2 &94.3 &47.0 &60.4 &45.9 &94.7 &95.8 &95.8 &100.0 \\
         
         
         Gemini-2.5-Pro \cite{comanici2025gemini2_5}
         &- &53.5 &46.0 &37.3 &68.7 &54.3 &61.9 &43.9 &47.4 &68.7 \\
         
         
         GPT-5 \cite{singh2025gpt5}
         &- &55.0 &53.3 &34.4 &73.3 &47.5 &63.7 &48.6 &50.2 &68.9 \\

         \midrule

        \rowcolor{gray!10!blue!5} 
        \multicolumn{11}{l}{\textit{Open-sourced General Models}}\\

         
         LLaVA-OneVision-72B \cite{li2024llavaonevision}
         &- &40.2 &43.5 &23.9 &57.6 &37.5 &42.5 &39.9 &32.5 &44.6 \\
         
         
         
         InternVL3-8B \cite{zhu2025internvl3}
         &- &42.1 &66.0 &34.8 &43.6 &47.5 &48.0 &39.3 &26.2 &31.3 \\
         
         
         Qwen2.5-VL-7B-Instruct \cite{Qwen2.5-VL}
         &- &31.4 &40.9 &14.8 &43.4 &10.7 &38.6 &40.1 &33.0 &29.8 \\
         
         Qwen3-VL-2B-Instruct \cite{Qwen3-VL}
         &- &50.3 &62.1 &40.2 &71.4 &49.7 &52.2 &42.0 &30.4 &54.5 \\
         
         Qwen3-VL-8B-Instruct \cite{Qwen3-VL}
         &- &57.9 &67.5 &47.0 &76.3 &61.9 &58.0 &50.9 &35.0 &66.3 \\

         
         \midrule

        \rowcolor{gray!10!blue!5} 
        \multicolumn{11}{l}{\textit{Spatial Intelligence Models}}\\

        SpatialLadder-3B \cite{li2025spatialladder}
        &\xmark &45.7 &63.5 &34.3 &61.7 &43.9 &45.4 &44.8 &35.6 &36.4\\

        Spatial-MLLM-4B \cite{wu2025spatialmllm}
        &\xmark &48.4 &65.3 &34.8 &63.1 &45.1 &41.3 &46.2 &33.5 &46.3\\

        VG-LLM-8B \cite{zheng2025vgllm}
        &\xmark &50.7 &67.9 &37.7 &58.6 &62.0 &46.6 &40.7 &32.4 &59.2 \\

        SpatialStack-4B \cite{zhang2026spatialstack}
        &\xmark &60.9 &69.2 &45.4 &63.0 &63.2 &57.9 &68.4 &40.2 &79.6 \\
        

        VLM-3R-7B \cite{fan2025vlm3r}
        &\xmark &60.9 &70.2 &49.4 &69.2 &67.1 &65.4 &80.5 &45.4 &40.1 \\
        
        SpaceMind-8B \cite{zhao2025spacemind}
        &\xmark &69.6 &73.3 &61.4 &77.3 &74.2 &67.2 &88.4 &44.3 &70.6 \\

         \midrule
        
        3DRS-7B \cite{huang20253drs}
        &\cmark &45.9 &68.7 &34.8 &53.6 &56.6 &40.9 &43.2 &30.4 &39.2 \\

        Cambrian-S-3B \cite{yang2025cambrians}
        &\cmark &57.3 &70.7 &40.6 &68.0 &46.3 &64.8 &61.9 &27.3 &78.8 \\
        
        Cambrian-S-7B \cite{yang2025cambrians}
        &\cmark &67.5 &73.2 &50.5 &74.9 &72.2 &71.1 &76.2 &41.8 &80.1 \\

        VST-3B-SFT \cite{yang2025vst}
        &\cmark &57.9 &69.3 &45.4 &71.8 &62.4 &59.0 &46.0 &38.7 &70.2 \\
        
        VST-7B-SFT \cite{yang2025vst}
        &\cmark &60.6 &72.0 &44.4 &74.3 &68.3 &59.7 &55.8 &44.9 &65.2 \\

         \midrule

        GeoVR-2B (\textbf{ours}) 
        &\cmark &69.1 &67.7 &54.5 &73.9 &72.3 &71.3 &80.7 &45.9 &86.7 \\

        GeoVR-4B (\textbf{ours}) 
        &\cmark &69.6 &65.6 &58.1 &75.5 &71.6 &70.0 &89.3 &41.2 &85.6 \\

        
        \bottomrule
    \end{tabular}
}
\caption{Performance comparisons on VSI-Bench. "w/o $\mathcal{E}_{3D}$" indicates models does not require a 3D foundation model during inference.}
\label{tab:vsi}
\vspace{-10pt}
\end{table*}

\begin{table}[h]
\centering
\resizebox{\linewidth}{!}{
\begin{tabular}{c | c c c c | c}
\toprule
\multirow{2}{*}{\textbf{Method}} & \multicolumn{4}{c|}{\textbf{ScanQA}} & \textbf{SQA3D} \\
\cmidrule{2-6}
 & B-4 & M & R & C & EM-1 \\
 

\midrule
\rowcolor{gray!15} \multicolumn{6}{l}{\textit{3D/2.5D-Input Models}} \\
3D-LLM \cite{hong20233dllm} 
& 12.0 & 14.5 & 35.7 & 69.4 & - \\

LL3DA \cite{chen2024ll3da} 
& 13.5 & 15.9 & 37.3 & 76.8 & - \\

ChatScene \cite{huang2024chatscene} 
& 14.3 & 18.0 & 41.6 & 87.7 & 54.6 \\

3D-LLaVA \cite{deng20253dllava} 
& 17.1 & 18.4 & 43.1 & 92.6 & 54.5 \\

Video-3D-LLM \cite{zheng2025video3dllm} 
& 16.4 & 20.0 & 49.3 & 102.1 & 58.6 \\

LLaVA-3D \cite{zhu2025llava3D} 
& 16.4 & 20.8 & 49.6 & 103.1 & 60.1 \\

\midrule
\rowcolor{gray!15} \multicolumn{6}{l}{\textit{Video-Input Models}} \\


Spatial-MLLM-4B \cite{wu2025spatialmllm} 
& 14.8 & 18.4 & 45.0 & 91.8 & 55.9 \\

GeoAlign-4B \cite{liu2026geoalign} 
&15.7 &19.4 &48.2 &99.4 &60.3 \\

3DRS-7B \cite{huang20253drs} 
& - & - & - & 104.8 & 60.6 \\

VLM-3R-7B \cite{fan2025vlm3r} 
& 15.5 & 19.7 & 49.1 & 101.9 & 60.7 \\

SpaceMind-8B \cite{zhao2025spacemind} 
& - & - & - & - & 61.1 \\

\midrule
GeoVR-2B (\textbf{ours})
& 15.3 & 20.7 & 49.3 & 103.3 & 61.4 \\

GeoVR-4B (\textbf{ours})
& 15.9 & 21.1 & 50.0 & 104.5 & 63.9 \\

\bottomrule
\end{tabular}
}
\caption{Performance comparisons on ScanQA and SQA3D.}
\label{tab:qa}
\vspace{-10pt}
\end{table}

\section{Experiments}
\label{sec:experiment}

\subsection{Implementation Details} 
\textbf{Backbone.} We adopt Qwen3-VL-2B/4B-Instruct \cite{Qwen3-VL} as the base model, VGGT-1B \cite{wang2025vggt} as the 3D foundation teacher, and DA3-Metric-Large \cite{lin2025depthanything3} as the metric depth model for real-world scale calibration. We also explore other 3D Foundation models, including VGGT-$\Omega$-1B \cite{wang2026vggtomega} and DepthAnything3-Giant \cite{lin2025depthanything3} as the 3D teacher in Sec. \ref{subsec:ablation}.

\textbf{Training Setup.} We train the model on a hybrid dataset comprising VSI-590K \cite{yang2025cambrians} and VLM-3R \cite{fan2025vlm3r} for 1 epoch. During training, 4 to 32 frames are sampled. The model is optimized using the AdamW optimizer with a global batch size of 32 and a learning rate of $2 \times 10^{-5}$. Specifically, the newly initialized tokens and auxiliary heads are optimized with a learning rate of $1 \times 10^{-4}$. Throughout the entire training process, both the 2D vision encoder and the auxiliary 3D teacher models are kept frozen. For the multi-scale geometric representation alignment, we extract hierarchical geometric features from the 5th, 12th, 18th, and 24th layers of VGGT as our distillation targets. 


\begin{table*}[!t]\large
    \centering
    \resizebox{\linewidth}{!}{
    \begin{tabular}{ c | c c c c | c | c c c c | c c c c }
        \toprule

        \multirow{2}{*}{\textbf{\#}}
        & \multirow{2}{*}{$\mathcal{L}_{cam}$}
        & \multirow{2}{*}{$\mathcal{L}_{depth}$}
        & \multirow{2}{*}{$\mathcal{L}_{scale}$} 
        & \multirow{2}{*}{$\mathcal{L}_{align}$} 
        & \multirow{2}{*}{\textbf{Avg.}} 
        & \multicolumn{4}{c|}{\cellcolor{orange!10}\textbf{Numerical Answer}} 
        & \multicolumn{4}{c}{\cellcolor{yellow!10}\textbf{Multiple-Choice Answer}} 
        \\

         \cmidrule(lr){7-10} \cmidrule(lr){11-14} 

         &
         &
         &
         &
         &
         &\rotatebox{0}{\textit{Obj. Count}} 
         &\rotatebox{0}{\textit{Abs. Dist}} 
         &\rotatebox{0}{\textit{Obj. Size}} 
         &\rotatebox{0}{\textit{Room Size}} 
         &\rotatebox{0}{\textit{Rel. Dis}}  
         &\rotatebox{0}{\textit{Rel. Dir}} 
         &\rotatebox{0}{\textit{Route Plan}} 
         &\rotatebox{0}{\textit{Appr. Order}}
         \\

        \midrule

         (0) &- &- &- &- 
         &56.7 &64.7 &39.4 &70.1 &48.8 &60.2 &57.7 &36.8 &76.7 \\

         
         



         


         (1) &\cmark &- &- &- 
         &59.8 &66.8 &40.2 &72.1 &60.5 &56.1 &66.9 &36.6 &79.1 \\
         
         (2) &- &\cmark &- &- 
         &59.7 &62.3 &40.5 &69.5 &62.5 &61.7 &66.4 &35.1 &79.3 \\

         (3) &\cmark &\cmark &- &- 
         &60.3 &65.5 &40.2 &72.0 &55.5 &60.6 &71.6 &39.7 &77.4 \\

         (4) &\cmark &\cmark &\cmark &- 
         &60.9 &68.1 &40.5 &72.7 &58.9 &58.6 &65.4 &43.3 &79.8 \\

         (5) &- &- &- &\cmark 
         &57.5 &63.6 &40.8 &69.6 &54.5 &57.6 &62.2 &35.8 &75.9 \\
         
         (6) &\cmark &\cmark &\cmark &\cmark 
         &62.1 &68.3 &42.5 &72.5 &62.5 &60.7 &66.6 &42.3 &81.2 \\
         
        \bottomrule
    \end{tabular}
}
\caption{Ablation study on Multi-task Geometric Learning, which shows
that simultaneous training with camera, depth, scale, and alignment yields the highest performance on VSI-Bench. ID \# (0) denotes the model finetuned with only $\mathcal{L}_{text}$.}
\label{tab:ablation_loss}
\vspace{-10pt}
\end{table*}

\begin{table}[!t]\Large
    \centering
    \resizebox{\linewidth}{!}{%
        \begin{tabular}{c|c|cccccccc}

        $\mathcal{E}_{3D}$
        &  \textbf{Avg.} 
        &\rotatebox{75}{\textit{Obj. Count}} 
        &\rotatebox{75}{\textit{Abs. Dist}} 
        &\rotatebox{75}{\textit{Obj. Size}} 
        &\rotatebox{75}{\textit{Room Size}} 
        &\rotatebox{75}{\textit{Rel. Dis}}  
        &\rotatebox{75}{\textit{Rel. Dir}} 
        &\rotatebox{75}{\textit{Route Plan}} 
        &\rotatebox{75}{\textit{Appr. Order}} \\

        &   
        &
        \multicolumn{4}{c}{\cellcolor{orange!10}\textbf{Numerical Answer}} &
        \multicolumn{4}{c}{\cellcolor{yellow!10}\textbf{Multiple-Choice Answer}} \\
        
        \toprule

        VGGT \cite{wang2025vggt} 
        & 62.1 & 68.3 & 42.5 & 72.5 & 62.5 & 60.7 & 66.6 & 42.3 & 81.2 \\

        VGGT-$\Omega$ \cite{wang2026vggtomega} 

        & 60.7 & 68.0 & 39.8 & 71.0 & 58.3 & 61.9 & 64.6 & 43.5 & 78.2 \\

        DA-3 \cite{lin2025depthanything3} 
        & 58.7 & 67.6 & 40.1 & 71.1 & 54.3 & 60.7 & 64.4 & 33.5 & 78.0 \\

        \bottomrule
    \end{tabular}
}
\caption{Ablation study on different 3D Foundation Models.}
\label{tab:ablation_3d}
\vspace{-10pt}
\end{table}

\subsection{Evaluation} 
\label{subsec:result}

\textbf{Comparison on VSI-Bench.}
We evaluate on VSI-Bench \cite{yang2025vsibench}, which contains over 5,000 QA pairs from egocentric videos in ScanNet, ScanNet++, and ARKitScenes. The benchmark includes Multiple-Choice (MCA) and Numerical Answer (NA) formats. As shown in Table \ref{tab:vsi}, our GeoVR models demonstrate superior spatial reasoning capabilities. GeoVR-2B and GeoVR-4B achieve impressive average scores of 69.1 and 69.6, respectively, yielding massive improvements over their corresponding baselines. Notably, they consistently surpass leading proprietary models like GPT-5 (55.0) and massive open-source generalists such as LLaVA-OneVision-72B (40.2). While dedicated spatial models like SpaceMind-8B achieve a competitive score of 69.6, they heavily rely on active 3D foundation models during inference, imposing a severe computational bottleneck. In contrast, GeoVR-4B matches this peak performance with absolutely zero additional architectural overhead and only half the parameters. Furthermore, among spatial models w/o $\mathcal{E}_{3D}$, both GeoVR variants easily outperform much larger competitors like Cambrian-S-7B (67.5). Detailed metric analysis highlights GeoVR-4B's exceptional physical grounding, particularly dominating in complex spatial layout and tracking tasks such as \textit{Abs. Dist} (58.1) and \textit{Route Plan} (89.3).

\textbf{Comparison on ScanQA and SQA3D.}
For ScanQA \cite{azuma2022scanqa} and SQA3D \cite{ma2022sqa3d} that require spatial reasoning in 3D scenes, we evaluate on their validation split using standard language metrics (CIDEr, BLEU, METEOR, ROUGE-L) and exact-match accuracy (EM). Following previous protocol \cite{fan2025vlm3r, wu2025spatialmllm}, we train the model on the training splits of ScanQA and SQA3D for a standardized evaluation setup. As shown in Table \ref{tab:qa}, our GeoVR models demonstrate remarkable spatial reasoning capabilities using purely 2D video inputs. Notably, the compact GeoVR-2B outperforms several recent 3D/2.5D-dependent models, such as LLaVA-3D \cite{zhu2025llava3D}, achieving higher CIDEr (103.3) and EM-1 (61.4) scores without requiring explicit point clouds during inference. Furthermore, GeoVR-4B establishes a new state-of-the-art among video-input models, recording a standout 63.9 EM-1 on SQA3D and significantly surpassing much larger baselines like VLM-3R-7B \cite{fan2025vlm3r} and SpaceMind-8B \cite{zhao2025spacemind}. These results strongly validate that our geometric representation restructuring successfully endows MLLMs with strong 3D scene comprehension while maintaining high parameter efficiency.

\subsection{In-Depth Analysis}
\label{subsec:ablation}
Unless otherwise specified, we establish our default experimental setting using Qwen3-VL-2B-Instruct as the MLLM and VGGT as the 3D teacher. All ablated models are only trained on the video subset of VSI-590K (around 374K samples) for 1 epoch, with a maximum of 8 frames per video.

\textbf{3D Foundation Model Backbone.}
We first investigate the impact of the 3D teacher model's capacity with three different $\mathcal{E}_{3D}$ backbones, including VGGT \cite{wang2025vggt}, VGGT-$\Omega$ \cite{wang2026vggtomega}, and DepthAnything-3 (DA-3) \cite{lin2025depthanything3}. For fair comparison, all models in this setting are jointly supervised by the full set of geometric targets. Specifically, for the multi-scale feature distillation, we extract representations from layers $\{5, 12, 18, 24\}$ for both VGGT and VGGT-$\Omega$, while layers $\{20, 28, 34, 40\}$ for DA-3. As in Table \ref{tab:ablation_3d}, the base VGGT surprisingly outperforms the stronger VGGT-$\Omega$ variant. We attribute this to the architectural design of VGGT-$\Omega$, which replaces a portion of its global attention with \textit{register attention} to reduce computational costs. While such an aggregated scene representation might be more efficient for certain downstream 3D reconstruction tasks, it inevitably compromises fine-grained spatial correspondences within dense image tokens, limiting the MLLM's ability to acquire robust geometric representations. Furthermore, both VGGT and VGGT-$\Omega$ models consistently surpass DA-3.

\textbf{Multi-task Geometric Learning.}
To validate the necessity of our multi-objective learning strategy, we conduct an ablation on the proposed geometric constraints from the 3D foundation model. As shown in Table \ref{tab:ablation_loss}, the baseline model (ID \# (0)) trained solely with text supervision achieves an average score of 56.7. Introducing only camera pose $\mathcal{L}_{cam}$ improves the performance to 59.8, notably boosting view-dependent metrics like \textit{Rel. Dir} (from 57.7 to 66.9). Conversely, applying only depth prediction $\mathcal{L}_{depth}$ raises the average to 59.7, with significant gains in metrics such as \textit{Room Size} (from 48.8 to 62.5). Combining them further elevates the average to 60.3, confirming that both tasks inject distinct yet complementary spatial awareness. The addition of metric scale calibration $\mathcal{L}_{scale}$ further raises the score to 60.9, proving its crucial role in helping the model understand absolute physical scales and distances in the real world. While applying geometric representation alignment $\mathcal{L}_{align}$ alone yields a modest gain (57.5), integrating all four geometric constraints achieves the highest overall performance (62.1). This demonstrates a strong complementary synergy: explicit geometric regressions provide rigid physical grounding, while implicit feature distillation ensures robust, multi-scale 3D representations.

\begin{table}[!t]\small
    \centering
    \begin{tabular}{c|c|c}
        \toprule
        
        $\boldsymbol{\mathcal{E}_{3D}}$ 
        & \textbf{Aligned Layer$^{th}$} 
        & \textbf{VSI-Bench} \\
        
        \midrule
        
        \multirow{6}{*}{\makecell{VGGT-$\Omega$ \cite{wang2026vggtomega} \\ ($L_{3D}$=24, $D_{3D}$=1024)}} 
        & 12 & 58.14 \\
        & 18 & 57.96 \\
        & 24 & 57.90 \\
        & \{12, 24\} & 57.25 \\
        & \{5, 18\} & 56.74 \\
        & \{5, 12, 18, 24\} & 59.67 \\
        
        \bottomrule
        
    \end{tabular}
    \caption{Ablation study on alignment strategy.}
    \label{tab:ablation_align}
    \vspace{-10pt}
\end{table}

\textbf{Alignment at Different Layers.}
We further investigate the effect of aligning different transformer layers of the MLLM to the 3D teacher. Here, we only keep the $\mathcal{L}_{align}$ loss active (discarding explicit geometric regressions $\mathcal{L}_{cam}$, $\mathcal{L}_{depth}$, and $\mathcal{L}_{scale}$) to strictly isolate the impact of pure feature-level distillation. As shown in Table \ref{tab:ablation_align}, when distilling from a single layer of VGGT-$\Omega$, aligning with the middle layer (the 12th) yields the best performance (58.14), slightly outperforming the deeper layers (the 18th and 24th). Interestingly, naively pairing two layers (e.g., [12, 24] or [5, 18]) leads to a noticeable performance drop, decreasing to 57.25 and 56.74, respectively. We attribute this to optimization conflicts caused by an incomplete hierarchical representation. However, when we apply a proportional, multi-scale alignment covering the entire backbone uniformly ([5, 12, 18, 24]), the performance surges to a peak of 59.67. This demonstrates that a comprehensive and evenly distributed distillation strategy is essential for the MLLM to progressively internalize 3D spatial priors, seamlessly bridging low-level geometry with high-level semantics.

\begin{figure}[t]
\begin{center}
\includegraphics[scale=0.66]{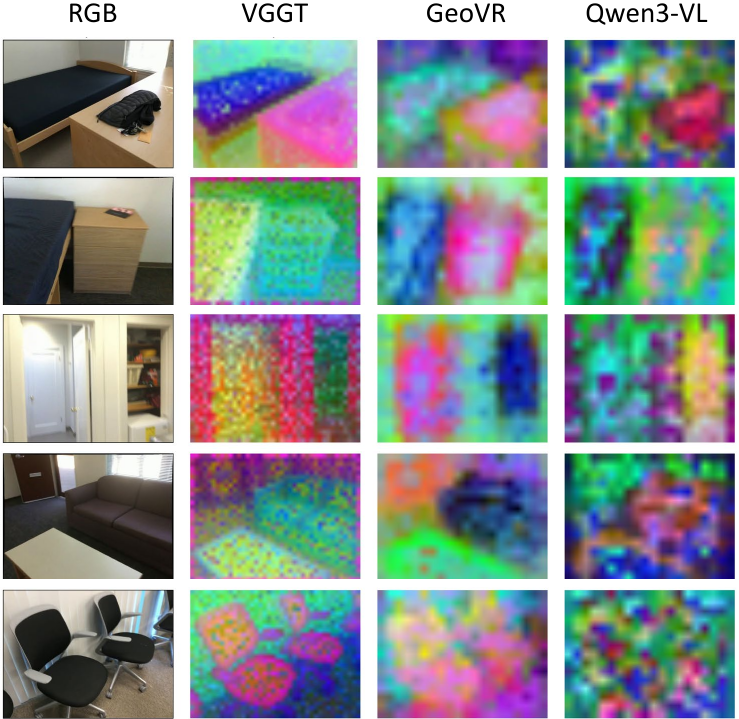}
\end{center}
\caption{PCA projections of visual representations.}
\label{fig:vis_pca}
\end{figure}

\begin{table}[!t]\small
    \centering
    \setlength{\tabcolsep}{8pt} 
    \begin{tabular}{lccc}
        \toprule
        \textbf{Depth Head} & \textbf{Params} & \textbf{L1 Loss} & \textbf{SILog Loss} \\
        \midrule
        MLP Head   & 13.6M & 58.42 & 59.31 \\
        DPT Head   & 32.7M & 58.48 & 58.87 \\
        Dense Head & 32.3M & 60.30 & 58.50 \\
        \bottomrule
    \end{tabular}
    \caption{Ablation study on depth prediction heads and loss.}
    \label{tab:ablation_depth}
    \vspace{-10pt}
\end{table}

\textbf{Depth Prediction Heads and Loss.} 
We evaluate how the architecture of the depth prediction heads influences learning. Under pure $\mathcal{L}_{depth}$ supervision from VGGT-$\Omega$, we try: (1) \textit{DPT Head}, which follows the exact dense vision transformer \cite{ranftl2021dpt} design used in VGGT, primarily composed of hierarchical convolutional blocks; (2) \textit{MLP Head}, a minimalist architecture consisting merely of a 3-layer MLP; and (3) \textit{Dense Head}, a hybrid design blending convolutions and MLPs. Additionally, we compare the L1-based loss in Eq. (\ref{equ:loss_depth}) against the scale-invariant logarithmic (SILog) loss \cite{eigen2014silog}. As shown in Table \ref{tab:ablation_depth}, while the SILog loss notably improves the lightweight MLP and DPT heads by relaxing the absolute scale penalty, the \textit{Dense Head} achieves the highest overall performance (60.30) when supervised by the L1 loss. Prioritizing absolute spatial reasoning accuracy over parameter efficiency, we adopt the \textit{Dense Head} with L1 supervision.

\begin{figure}[t]
\begin{center}
\includegraphics[scale=0.65]{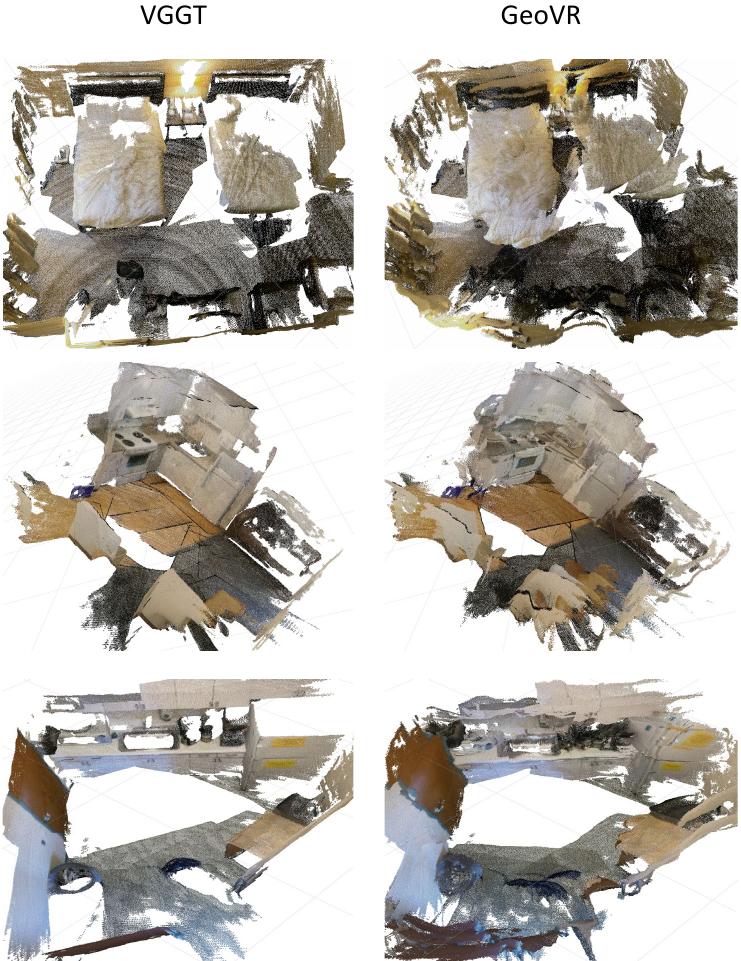}
\end{center}
\caption{3D point clouds reconstructed from 2D videos.}
\label{fig:vis_3d}
\vspace{-10pt}
\end{figure}

\textbf{Feature Visualization.}
To qualitatively demonstrate the effectiveness of our geometric representation restructuring, we visualize the internal feature representations and the reconstructed 3D scenes. In Fig. \ref{fig:vis_pca}, we project the high-dimensional visual tokens into RGB space using PCA. The original MLLM (Qwen3-VL) exhibits noisy and geometrically inconsistent representations, failing to delineate clear object boundaries or spatial layouts across different views. In contrast, after our multi-objective geometric learning, the representations of GeoVR become more structured and smoother, maintaining sharper geometric consistency that closely mirrors the explicit multi-view representations of the 3D teacher (VGGT). Furthermore, in Fig. \ref{fig:vis_3d}, we leverage the predicted depth maps and camera poses from GeoVR to reconstruct the scene by directly unprojecting the 2D video pixels into 3D point clouds. The visualizations confirm that GeoVR can kind of recover 3D scene structures and spatial layouts, demonstrating a level of spatial fidelity comparable to the 3D foundation model. This strongly supports the conclusion that our method helps MLLM effectively internalize the physical 3D world solely from 2D observations.

\section{Conclusion}
In this paper, we introduce GeoVR, a novel framework designed to awaken spatial intelligence within MLLMs relying purely on 2D video sequences. We propose a multi-objective geometric learning paradigm. By estimating inter-frame camera poses, regressing dense depth maps, calibrating real-world metric scales, and distilling multi-scale geometric priors from a pre-trained 3D foundation teacher, GeoVR fundamentally restructures the MLLM's internal semantic latent space into geometry-aware representations. Extensive experiments demonstrate that our method significantly enhances the model's capabilities in spatial reasoning. In the future, we plan to scale the GeoVR paradigm to larger MLLM architectures and datasets and explore its potential in more complex spatial intelligence tasks.

{
    \small
    \bibliographystyle{ieeenat_fullname}
    \bibliography{main}
}


\end{document}